\documentclass[times,twocolumn,final]{elsarticle}

\usepackage{framed,multirow}

\usepackage{amssymb}
\usepackage{latexsym}

\usepackage{url}
\usepackage{xcolor}
\definecolor{newcolor}{rgb}{.8,.349,.1}

\usepackage{xr}

\usepackage{times}
\usepackage{graphicx} 
\usepackage{epstopdf}
\usepackage{subcaption}
\usepackage{amsmath}
\DeclareMathOperator*{\argmin}{\arg\!\min}
\usepackage{natbib}

\usepackage{soul}

\usepackage{algorithm}
\usepackage{algorithmic}

\usepackage{amsthm}

\usepackage{lineno,hyperref}

\begin{document}


\title{On denoising autoencoders trained to minimise binary cross-entropy}

\author[1]{Antonia Creswell}
\author[1]{Kai Arulkumaran}
\author[1,2]{Anil A. Bharath}

\address[1]{Department of Bioengineering, Faculty of Engineering, Imperial College London, London SW7 2AZ, United Kingdom}
\address[2]{Data Science Institute, Imperial College London, London SW7 2AZ, United Kingdom}

\maketitle

\begin{abstract} 
Denoising autoencoders (DAEs) are powerful deep learning models used for feature extraction, data generation and network pre-training. DAEs consist of an encoder and decoder which may be trained simultaneously to minimise a loss (function) between an input and the reconstruction of a corrupted version of the input. There are two common loss functions used for training autoencoders, these include the mean-squared error (MSE) and the binary cross-entropy (BCE). When training autoencoders on image data a natural choice of loss function is BCE, since pixel values may be normalised to take values in $[0,1]$ and the decoder model may be designed to generate samples that take values in $(0,1)$.
We show theoretically that DAEs trained to minimise BCE may be used to take gradient steps in the data space towards regions of high probability under the data-generating distribution. Previously this had only been shown for DAEs trained using MSE. As a consequence of the theory, iterative application of a trained DAE moves a data sample from regions of low probability to regions of higher probability under the data-generating distribution. Firstly, we validate the theory by showing that novel data samples, consistent with the training data, may be synthesised when the initial data samples are random noise. Secondly, we motivate the theory by showing that initial data samples synthesised via other methods may be improved via iterative application of a trained DAE to those initial samples.




\end{abstract}



\section{Introduction}

Autoencoders are a class of neural network models that are conceptually simple, yet provide a powerful means of learning \emph{useful representations} of observed data through unsupervised learning \cite{bengio2009learning}. Trained autoencoders can be used in downstream tasks, either through extracting the learned representations as input for other algorithms \cite{lange2012autonomous}, or by fine-tuning the existing model for other tasks \cite{vincent2008extracting}. Additionally, recent work has led to the development of generative autoencoders, i.e., autoencoders that both learn useful representations and allow the synthesis of novel data samples that are consistent with the training data \cite{kingma2013auto,rezende2014stochastic,makhzani2015adversarial}.

Autoencoders consist of two models, an encoder and a decoder, arranged in series. The two models may be trained simultaneously to minimise a reconstruction loss between the input to the encoder, and the output of the decoder. A common variant of the autoencoder is the denoising autoencoder (DAE), which is trained to recover clean versions of corrupted input samples \cite{vincent2008extracting, vincent2010stacked}. 

Formally, a DAE with encoder, $e(\cdot; \phi)$, decoder, $d(\cdot;\theta)$ and corruption process, $C(\cdot)$, is trained to minimise:

\begin{equation}
    \mathbb{E}_X \ell(x, d(e(C(x);\phi);\theta)),
\end{equation}

where $x \in X$ are data samples whose underlying distribution is $p(X)$. Parameters $\phi$ and $\theta$ are learned. There may be additional regularisation terms in the loss function.

It has been shown \cite{alain2014regularized} that the optimal reconstruction function, $R^*(\cdot)$ learned by a DAE trained to minimise the \textbf{mean-squared} reconstruction error approximates the gradient of the log-likelihood of the data-generating distribution with respect to the data. This theoretical result has been a driving force behind many publications \cite{nguyen2016plug, rasmus2015semi,huszar2017variational} (see Section \ref{sec:previous} for more details). 

However, the mean-squared error (MSE) is only one of two common loss functions used when training autoencoders. This is especially true when the data distribution consists of images which take values in $[0, 1]$ and the pixel intensity can be thought of as the probability of a pixel being ``on''. In this situation, it is possible to use the binary cross-entropy (BCE) loss, which naturally applies to random variables that may be ``on'' or ``off''.

In this paper, we extend the theory presented by Alain and Bengio \cite{alain2014regularized} (Section \ref{sec:theory}) and present empirical results to validate the theory in a practical setting (Section \ref{sec:valid}). Furthermore, we present an application (Section \ref{sec:application}), where by applying the theory we may improve samples synthesised from both variational autoencoders \cite{kingma2013auto, rezende2014stochastic} and adversarial autoencoders \cite{makhzani2015adversarial} trained with the denoising criterion.

\section{Background}

Alain and Bengio \cite{alain2014regularized} show that if a regularised DAE is trained to minimise the reconstruction loss:

\begin{equation}
\label{eqn:MSE}
\theta^*, \phi^* = \argmin_{\theta, \phi} \mathbb{E}_X[|| d(e(\tilde{x}; \phi);\theta) - x ||_2^2]
\end{equation}

where $\tilde{x} = C(x) = x + \epsilon$, where $\epsilon \sim \mathcal{N}(0, \mathbb{I}\sigma^2)$, the optimal reconstruction function, $R^*_\sigma = d(e(x; \phi^*);\theta^*)$, 
in the limit, as $\sigma \rightarrow 0$, the optimal reconstruction function is given by \cite{alain2014regularized}:

\begin{equation}
\label{eqn:limit}
\lim_{\sigma \rightarrow 0} \{R^*_\sigma(x)\} = x + \sigma^2 \frac{\partial \log p(x)}{\partial x}.
\end{equation}

Equation (\ref{eqn:limit}) is equivalent to a gradient ascent step in data space, $X$, moving towards regions of higher likelihood, where the step size is given by $\sigma^2$.


\section{Theory}
\label{sec:theory}

Alain and Bengio \cite{alain2014regularized} only showed that Equation (\ref{eqn:limit}) held for the MSE loss function given in Equation (\ref{eqn:MSE}). Here we show that Equation (\ref{eqn:limit}) also holds when BCE is used for the loss function rather than MSE.

Consider a reconstruction model, $R(\tilde{x}) = d(e(\tilde{x}; \phi);\theta)$, trained to minimise:

\begin{equation}
\mathcal{L}_{DAE} = \mathbb{E}_X [\ell_\text{BCE}(x,R(\tilde{x}))],
\end{equation}
where $\ell_\text{BCE}$ is the BCE loss; this expands to:
\begin{equation}
\begin{split}
& \mathcal{L}_{DAE} \\ & = - \int_X p(x) \mathbb{E}_{\epsilon \sim \mathcal{N}(0, \mathbb{I}\sigma^2)} [x \log(R(\tilde{x})) \\ & + (1- x) \log (1 - R(\tilde{x}))] dx.
\end{split}
\end{equation}
Letting $\tilde{x} = x + \epsilon$ and shortening $\mathbb{E}_{\epsilon \sim \mathcal{N}(0, \mathbb{I}\sigma^2)}$ to $\mathbb{E}_\epsilon$ for ease of notation:
\begin{equation}
\begin{split}
& \mathcal{L}_{DAE}  = \\ & - \int_X p(\tilde{x} - \epsilon) \mathbb{E}_\epsilon [(\tilde{x} - \epsilon) \log(R(\tilde{x})) \\ & + (1 - \tilde{x} + \epsilon) \log (1 - R(\tilde{x}))]dx.
\end{split}
\end{equation}
The optimal reconstruction model, $R^*_\sigma$ is obtained by solving the following equation:
\begin{equation}
\label{eqn:solve}
\frac{\partial \mathcal{L}_\text{DAE}}{\partial R(\tilde{x})} \Bigg|_{R(\tilde{x}) = R^*_\sigma(\tilde{x})}=0,
\end{equation}
where,
\begin{equation}
\label{eqn:derivative}
\frac{\partial \mathcal{L}_\text{DAE}}{\partial R(\tilde{x})} = - \mathbb{E}_\epsilon p(\tilde{x}-\epsilon) \Bigg[\frac{(\tilde{x}-\epsilon)}{R(\tilde{x})} + \frac{- (1 - \tilde{x} +\epsilon)}{(1 - R(\tilde{x}))} \Bigg].
\end{equation}
Substituting Equation (\ref{eqn:derivative}) into (\ref{eqn:solve}) and rearranging we get:
\begin{equation}
\label{eqn:BCE_opt}
R^*_\sigma(x)= \frac{\mathbb{E}_\epsilon[p(x-\epsilon)(x-\epsilon)] }{\mathbb{E}_\epsilon [p(x-\epsilon)]}
\end{equation}

which is the same the intermediate result obtained by Alain and Bengio \cite{alain2014regularized} in their derivation of Equation (\ref{eqn:limit}). Therefore Equation (\ref{eqn:limit}) also holds for DAEs trained using BCE.

\section{Experiments}

The purpose of our experiments is twofold. To validate the theory presented in Section \ref{sec:theory}, we show that it is possible to synthesise novel data samples that are consistent with the training data, starting from random noise. We then provide a practical motivation, by showing that samples synthesised via generative autoencoders \cite{kingma2013auto,rezende2014stochastic,makhzani2015adversarial} using existing sampling methods can be improved via iterative application of a trained denoising generative autoencoder to those initial samples.

\subsection{Setup}

For our experiments we train denoising variants of two state-of-the-art generative autoencoder models---the variational autoencoder (VAE) \cite{kingma2013auto,rezende2014stochastic} and the adversarial autoencoder (AAE) \cite{makhzani2015adversarial}---for the two reasons specified above. Firstly, we use these models in order to validate Equation (\ref{eqn:limit}) (which we have previously shown is a consequence of Equation (\ref{eqn:BCE_opt})). Secondly, we demonstrate how our sampling method, as detailed in Equation (\ref{eqn:iter}), may be used to improve the perceptual quality of data samples synthesised via other methods.

Generative autoencoders are autoencoders which are trained with a reconstruction loss, as well as an extra regularisation loss which encourages the distribution of the encoded samples to conform to a specified prior distribution; for simplicity this is often chosen to be the multivariate normal distribution $\mathcal{N}(0, \mathbb{I})$, and we use this standard choice in our own experiments. For details of the derivation and implementation of these regularisation losses, we refer readers to the original literature \cite{kingma2013auto,makhzani2015adversarial}.

We turn these models into denoising variants---denoising variational autoencoders (DVAEs) \cite{im2015denoising} and denoising adversarial autoencoders (DAAEs)---simply by replacing the reconstruction loss with a denoising loss. Following Alain and Bengio \cite{alain2014regularized}, we use an additive Gaussian noise corruption process; we use noise sampled from $\mathcal{N}(0, 0.25\mathbb{I})$. Each of our denoising generative autoencoder models is trained till convergence, such that the learned reconstruction function, $\hat{R}^*_\sigma(\cdot)$ is close to the optimal $R^*_\sigma(\cdot)$.

In line with the recommendations of Radford et al. \cite{radford2015unsupervised}, we use strided convolutional layers in the encoders, followed by fractionally-strided convolutions in the decoders. Each layer is followed by batch normalisation \cite{ioffe2015batch} and ReLU nonlinearities, except for the final layer of the decoders, which have their values restricted to $(0, 1)$ by a sigmoid function. The architecture of the encoders and decoders match the discriminator and generator networks of Radford et al. \cite{radford2015unsupervised}. For the adversarial model in the DAAE, we use a fully-connected network with dropout and leaky ReLU nonlinearities. When we perform sampling using our trained generative autoencoders, we utilise minibatch statistics for the batch normalisation layers, as it can help compensate for inputs that are far from the training distribution.

For training we use the CelebA dataset, which consists of two hundred thousand aligned and cropped images of celebrity faces \cite{liu2015deep}. It is widely used for the qualitative evaluation of generative models as it contains a large amount of high quality image data, and the use of human faces makes it easier for unrealistic qualities of synthesised samples to be spotted. We preprocess the images by cropping and resizing them to $64\times64$px, and then train our models using the Adam optimiser (using hyperparameters $\alpha = 0.0002$, $\beta_1 = 0.5$ and $\beta_2 = 0.999$) \cite{kingma2014adam} on the training split of the dataset for 20 epochs. All of our experiments were carried out using the Torch library \cite{collobert2011torch7}.

\subsection{Sampling From Noise}
\label{sec:valid}

In Section \ref{sec:theory} we showed that utilising an optimally trained reconstruction model is equivalent to a gradient step in the direction of more likely data samples. We can validate this empirically by iteratively applying a trained reconstruction model to an initial image, $x_0$ where each pixel is sampled independently from some random distribution. The trained reconstruction model is applied such that:
\begin{equation}
\label{eqn:iter}
x_{t+1} = d(e(x;\hat{\phi});\hat{\theta})=\hat{R}^*_\sigma(x_t),
\end{equation}
where $t>0$. 
According to Equation (\ref{eqn:limit}), each iteration should produce image samples that are more likely under the data-generating distribution. Qualitatively, this means that several iterations of applying the reconstruction model to random noise results in progressively more realistic faces.

\begin{figure}
  \captionsetup[subfigure]{justification=centering}
  \begin{subfigure}[]{0.32\linewidth}
    \includegraphics[width=\linewidth]{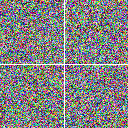}
    \subcaption{Initial samples\\(noise)}
  \end{subfigure}
  \hfill
  \begin{subfigure}[]{0.32\linewidth}
    \includegraphics[width=\linewidth]{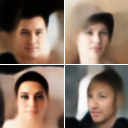}
    \subcaption{DVAE\\ (10 steps)}
  \end{subfigure}
  \hfill
  \begin{subfigure}[]{0.32\linewidth}
    \includegraphics[width=\linewidth]{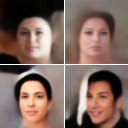}
    \subcaption{DVAE\\ (20 steps)}
  \end{subfigure}
  \\[1.5ex]
  \begin{subfigure}[]{0.32\linewidth}
    \includegraphics[width=\linewidth]{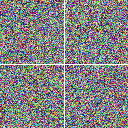}
    \subcaption{Initial samples\\(noise)}
  \end{subfigure}
  \hfill
  \begin{subfigure}[]{0.32\linewidth}
    \includegraphics[width=\linewidth]{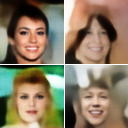}
    \subcaption{DAAE\\ (10 steps)}
  \end{subfigure}
  \hfill
  \begin{subfigure}[]{0.32\linewidth}
    \includegraphics[width=\linewidth]{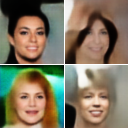}
    \subcaption{DAAE\\ (20 steps)}
  \end{subfigure}
  \caption{Applying the sampling procedure from Equation (\ref{eqn:iter}) to initial data samples drawn from a uniform distribution, $x_0 \sim U(0, 1)$, shown in (a, d), to produce samples $x_{10}$ (b, e) and $x_{20}$ (c, f). This shows the clear emergence of samples consistent with the training data, starting from noise. For (b, c) the reconstruction function is a trained DVAE, and for (e, f) it is a trained DAAE.}
  
  \label{no_noise_samples}
\end{figure}

The results of our sampling procedure, given by Equation (\ref{eqn:iter}), are shown in Figure (\ref{no_noise_samples}). The initial samples are noise drawn from a uniform distribution and the proceeding samples appear more like faces. A few reconstruction iterations are sufficient to produce face-like images, and several further iterations result in a significant improvement in perceptual quality. These results show that iterative application of Equation (\ref{eqn:iter}) moves samples from regions of low probability under the data-generating distribution to regions of higher probability.

\begin{figure*}
  \captionsetup[subfigure]{justification=centering}
  \begin{subfigure}[]{0.15\linewidth}
    \includegraphics[width=\linewidth]{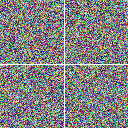}
    \subcaption{Initial samples\\(noise)}
  \end{subfigure}
  \hfill
  \begin{subfigure}[]{0.15\linewidth}
    \includegraphics[width=\linewidth]{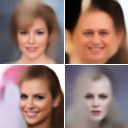}
    \subcaption{DVAE\\ (10 steps)}
  \end{subfigure}
  \hfill
  \begin{subfigure}[]{0.15\linewidth}
    \includegraphics[width=\linewidth]{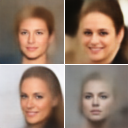}
    \subcaption{DVAE\\ (20 steps)}
  \end{subfigure}
  \hfill
  \begin{subfigure}[]{0.15\linewidth}
    \includegraphics[width=\linewidth]{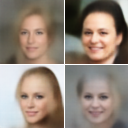}
    \subcaption{DVAE\\ (30 steps)}
  \end{subfigure}
  \hfill
  \begin{subfigure}[]{0.15\linewidth}
    \includegraphics[width=\linewidth]{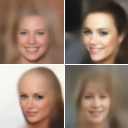}
    \subcaption{DVAE\\ (40 steps)}
  \end{subfigure}
  \hfill
  \begin{subfigure}[]{0.15\linewidth}
    \includegraphics[width=\linewidth]{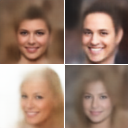}
    \subcaption{DVAE\\ (50 steps)}
  \end{subfigure}
  \\[1.5ex]
  \begin{subfigure}[]{0.15\linewidth}
    \includegraphics[width=\linewidth]{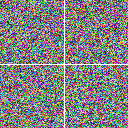}
    \subcaption{Initial samples\\(noise)}
  \end{subfigure}
  \hfill
  \begin{subfigure}[]{0.15\linewidth}
    \includegraphics[width=\linewidth]{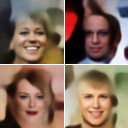}
    \subcaption{DAAE\\ (10 steps)}
  \end{subfigure}
  \hfill
  \begin{subfigure}[]{0.15\linewidth}
    \includegraphics[width=\linewidth]{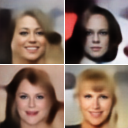}
    \subcaption{DAAE\\ (20 steps)}
  \end{subfigure}
  \hfill
  \begin{subfigure}[]{0.15\linewidth}
    \includegraphics[width=\linewidth]{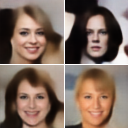}
    \subcaption{DAAE\\ (30 steps)}
  \end{subfigure}
  \hfill
  \begin{subfigure}[]{0.15\linewidth}
    \includegraphics[width=\linewidth]{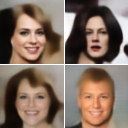}
    \subcaption{DAAE\\ (40 steps)}
  \end{subfigure}
  \hfill
  \begin{subfigure}[]{0.15\linewidth}
    \includegraphics[width=\linewidth]{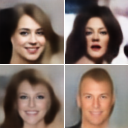}
    \subcaption{DAAE\\ (50 steps)}
  \end{subfigure}
  \caption{Starting from a random data sample drawn from a uniform distribution, $x_0 \sim U(0, 1)$ (a, g), and iteratively encoding and decoding using a trained DVAE (b-f) and a trained DAAE (h-l). Gaussian noise $\sim N(0, 0.25\mathbb{I})$ is added before each encoding step. By applying the sampling method from Equation (\ref{eqn:iter}), we observe samples moving along a manifold towards more likely regions. By adding noise, and having a fixed step size $\sigma^2$, (Equation (\ref{eqn:limit})) we may also see transitions where samples move from a region near lower a local maximum to a region closer to a higher local maximum (see Section \ref{sec:discussion} for more details).}
  \label{noise_samples}
\end{figure*}

If we consider that the support for images of valid faces exist on manifolds with lower dimensions than the data \cite{arjovsky2017towards}, then the gradient for samples far from the manifold may be very weak. One potential solution for this problem is to add a small amount of noise before each iteration (of applying Equation \ref{eqn:iter}), which has the effect of smoothing out the (log) probability distribution, hence making it easier to take gradient steps. We demonstrate the result of applying this technique in Figure \ref{noise_samples}, where we show the results of applying Equation  (\ref{eqn:iter}) for many iterations.


\subsection{Improving Initial Samples}
\label{sec:application}

In this section we show how the iterative sampling process described by Equation (\ref{eqn:iter}) can improve samples drawn from a DVAE or DAAE by first drawing samples in the standard way \cite{kingma2013auto}, and then applying iterative encoding and decoding to move the sample towards more likely regions of the data-generating distribution.

Normally, samples are drawn from a generative autoencoder by drawing a latent encoding, $z$ from the chosen prior distribution, and then passing it through the trained decoder. We refer to a data sample drawn in this way as $x_0$. In order to improve these data samples we propose iteratively applying a trained reconstruction function (Equation (\ref{eqn:iter})) in order to move the sample towards regions of higher probability under the data-generating distribution. The process is detailed in Equation (\ref{eqn:improve}), where we initialise the process with the conventional prior distribution $\mathcal{N}(0, \mathbb{I})$.

\begin{equation}
\label{eqn:improve}
\begin{split}
    z & \sim \mathcal{N}(0, \mathbb{I}) \\
    x_0 & = d(z, \hat{\phi})\\
    x_{t+1} & = \hat{R}^*_\sigma(x_t)
\end{split}
\end{equation}

\begin{figure}[H]
  \captionsetup[subfigure]{justification=centering}
  \begin{subfigure}[]{0.32\linewidth}
    \includegraphics[width=\linewidth]{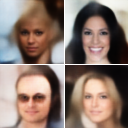}
    \subcaption{Initial samples\\ (DVAE)}
  \end{subfigure}
  \hfill
  \begin{subfigure}[]{0.32\linewidth}
    \includegraphics[width=\linewidth]{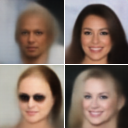}
    \subcaption{DVAE\\ (10 steps)}
  \end{subfigure}
  \hfill
  \begin{subfigure}[]{0.32\linewidth}
    \includegraphics[width=\linewidth]{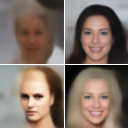}
    \subcaption{DVAE\\ (20 steps)}
  \end{subfigure}
  \\[1.5ex]
  \begin{subfigure}[]{0.32\linewidth}
    \includegraphics[width=\linewidth]{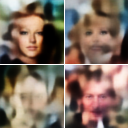}
    \subcaption{Initial samples\\ (DAAE)}
  \end{subfigure}
  \hfill
  \begin{subfigure}[]{0.32\linewidth}
    \includegraphics[width=\linewidth]{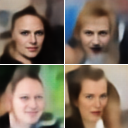}
    \subcaption{DAAE\\ (10 steps)}
  \end{subfigure}
  \hfill
  \begin{subfigure}[]{0.32\linewidth}
    \includegraphics[width=\linewidth]{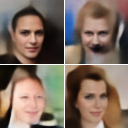}
    \subcaption{DAAE\\ (20 steps)}
  \end{subfigure}
  \caption{Samples are synthesised via the sampling process detailed in Equation (\ref{eqn:improve}). An initial sample $x_0$ is drawn from a trained DVAE (a) and DAAE (d). (a) and (d) are iteratively encoded and decoded using the same models---the DVAE for (b, c) and the DAAE for (e, f). Gaussian noise $\sim N(0, 0.25\mathbb{I})$ is added before each encoding step.}
  \label{improve_samples}
\end{figure}

The results of this procedure are shown in Figure \ref{improve_samples}. Although the quality of the initial samples vary greatly, we usually observe an improvement with several steps of the sampling procedure. This improvement is more noticeable when the initial samples are worse (see Figure \ref{improve_samples}(d, e, f)).

\section{Relation to Previous Work}
\label{sec:previous}



We consider our work in relation to previous work where denoising autoencoders have been used to learn both homogeneous transition operators of Markov chains \cite{nguyen2016plug, bengio2013generalized, bengio2014deep} and non-homogeneous transition operators \cite{salimans2015markov,bachman2015variational}. 

Bengio et al. \cite{bengio2013generalized, bengio2014deep} construct the transition operator of a Markov chain whose stationary distribution is the data-generating distribution. The transition operator is implemented by corrupting a data sample and reconstructing it. The reconstruction function is a trained DAE. Bengio et al. \cite{bengio2013generalized, bengio2014deep} initialise their sampling process with a sample that is consistent with the training data and apply their iterative sampling process to the MNIST \cite{lecun1998gradient} hand-written numbers dataset and  produce a series of samples from different modes in the data-generating distribution. For example, starting with a \textbf{9}, the process may generate \textbf{7}'s, \textbf{6}'s and \textbf{0}'s, as well as samples that do not correspond to numbers. Their experiments demonstrate that iterative sampling leads to convergence to a distribution, rather than a fixed point. Note that in the work of Bengio et al. \cite{bengio2013generalized, bengio2014deep}, transitions between modes are made possible by addition of sufficient noise before each reconstruction step.

In contrast to the work of Bengio et al. \cite{bengio2013generalized, bengio2014deep}, our work focuses on moving towards to a single data point in a distribution rather than movement between all modes in the data-generating distribution. Specifically, in our work, the addition of noise between iterations was used to encourage a smoother data space to take steps in (contrast Figures \ref{no_noise_samples} and \ref{noise_samples}), whereas in Bengio et al. \cite{bengio2013generalized, bengio2014deep} the corruption process was required to allow successful transition between modes. Note, that by applying the iterative process in Equation (\ref{eqn:iter}), starting from different initial random noise inputs, we are able to generate samples from different modes in the data-generating distribution.




Interestingly, Bengio et al. \cite{bengio2014deep, bengio2013generalized} found that their sampling process often lead to a series of nonsensical samples when transitioning between modes; they introduced the ``walkback" algorithm which involves updating the reconstruction model while sampling from it at the same time. The need for such a process indicates the drawbacks of attempting to transition between modes of a data distribution, especially when that data distribution consists of images.

When considering the specific application of DAEs to images, it may make more sense to consider a gradient method that searches for a single, highly likely point in a distribution, rather than aiming to capture the whole distribution starting from a single sample. This is because image samples exist on manifolds which are often in lower dimensions than the data space \cite{arjovsky2017towards}. This means that manifolds are unlikely to intercept \cite{arjovsky2017towards}, making it difficult to move from one mode or manifold to another while generating sampling that are consistent with the training data.


Other related work includes that of Nguyen et al. \cite{nguyen2015deep}, who define a Markov chain with a transition operation defined as:

\[x_{t+1} = x_t + \epsilon_{1,2} \nabla \log p(x_t) + \mathcal{N}(0, \epsilon_3^2), \]

where $\epsilon_{1,2}$ and $\epsilon_3$ are hyperparameters.
Nguyen et al. approximate $\nabla \log p(x_t) \approx (R(x_t)-x_t)$ and choose the step size to be $\epsilon_{1,2}$, rather than allowing $\sigma^2$ to be the step size, as we do with our transition operator in Equation (\ref{eqn:iter}). Nguyen et al. \cite{nguyen2015deep} train their DAE models using MSE because of the properties shown by Alain and Bengio \cite{alain2014regularized} for this specific cost function. Given our results, their models could be constructed using a BCE loss, which may be more appropriate for image data.

There are several other studies, where Markov chain sampling has been applied both during training and sample synthesis \cite{salimans2015markov, bachman2015variational} (and the ``walkback" algorithm of Bengio et al. \cite{bengio2013generalized}) however, we emphasise that our work has focused on applying Markov chains only during sample synthesis and not during training. Application during training may be another avenue for future work.

\section{Discussion}
\label{sec:discussion}

\begin{figure}[h]
  \centering
  \includegraphics[width=\columnwidth]{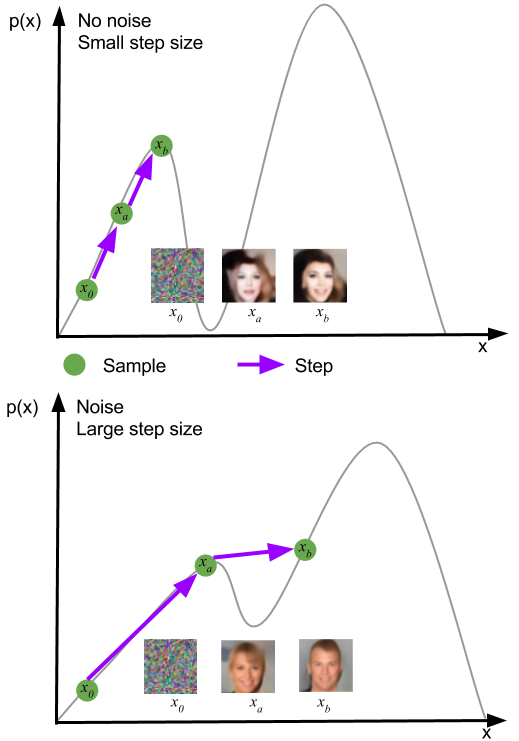}
  \caption{Gradient ascent towards regions of high probability under the data-generating distribution. Small steps are more likely to improve the likelihood of samples whilst preserving qualities such as identity. In contrast, the addition of noise before encoding can result in larger steps, and hence makes transitions between modes more likely.}
  \label{fig:diagram}
\end{figure}

Rather than trying to synthesise samples from all modes in the data-generating distribution \cite{bengio2013generalized}, our aim is to move towards more likely data samples. This may involve moving along manifolds towards more likely regions ---leading to small changes in identity, pose and even gender. Further, since we are taking gradient steps with fixed size, $\sigma^2$ (Equation (\ref{eqn:limit})), it is possible that if $\sigma$ is not sufficiently small we may ``overshoot'' a local maximum, and start gradient ascent on a different, higher, local maximum (see Figure \ref{fig:diagram}). This explains why there may be changes in identity, pose or gender between samples.

To produce a variety of samples---visiting different modes in the data-generating distribution---we may initialise the sampling process with samples from different location in the data generating distribution.

\section{Conclusion}

Building on previous work \cite{alain2014regularized}, we show that denoising autoencoders trained to minimise the binary cross-entropy loss may be used to approximate the gradient of the log density distribution of the data-generating distribution with respect to data samples. As a result, the sampling process, detailed in Equation (\ref{eqn:iter}), can be applied to any kind of autoencoder trained with a binary cross-entropy denoising loss.  Empirically, we validate our findings by showing that it is possible to synthesise novel samples consistent with the data-generating distribution starting from random noise. In addition, we provide a practical application, demonstrating that it is possible to improve the perceptual quality of initial samples drawn from denoising variants of variational \cite{kingma2013auto, rezende2014stochastic} and adversarial \cite{makhzani2015adversarial} autoencoders via our proposed sampling procedure.

\section*{Acknowledgements}

We would like to acknowledge the EPSRC for funding through a Doctoral Training studentship and the support of the EPSRC CDT in Neurotechnology.

\bibliography{paper}
\bibliographystyle{elsarticle-num}

\end{document}